\documentclass[sigconf]{acmart}

\usepackage[table]{xcolor}
\usepackage{booktabs}      
\usepackage{array}         
\usepackage{makecell}     
\usepackage{multirow}      

\usepackage{graphicx}      

\usepackage{pifont}        
\usepackage[T1]{fontenc}   
\usepackage[utf8]{inputenc}
\usepackage{microtype}   

\usepackage{amsmath}

\definecolor{lightgray}{gray}{0.9}

\newcommand{\thickhline}{\noalign{\hrule height 1pt}}

\copyrightyear{2026}
\acmYear{2026}
\setcopyright{cc}
\setcctype{by}
\acmConference[MM '26]{Proceedings of the 34th ACM International Conference on Multimedia}{November 10--14, 2026}{Rio de Janeiro, Brazil}
\acmBooktitle{Proceedings of the 34th ACM International Conference on Multimedia (MM '26), November 10--14, 2026, Rio de Janeiro, Brazil}
\acmDOI{10.1145/3767308.3836472}
\acmISBN{979-8-4007-2213-4/2026/11}
\begin{document}

\title{Vision-Grounded Semantic Learning and Alignment for Scalable Multilingual Text-Based Person Retrieval}

\author{Xinyu Zhou}
\email{xyzhou25@stu.suda.edu.cn}
\orcid{0009-0008-7547-1786}
\affiliation{%
  \institution{Soochow University}
  \city{Suzhou}
  \country{China}
}

\author{Fan Li}
\affiliation{%
  \institution{Soochow University}
  \city{Suzhou}
  \country{China}
}

\author{Shasha Chen}
\affiliation{%
  \institution{Soochow University}
  \city{Suzhou}
  \country{China}
}

\author{Min Cao}
\email{mcao@suda.edu.cn}
\authornote{Corresponding Author}
\affiliation{%
  \institution{Soochow University}
  \city{Suzhou}
  \country{China}
}

\author{Ahtamjan Ahmat}
\affiliation{%
 \institution{The Xinjiang Technical Institute of Physics \& Chemistry, Chinese Academy of Sciences}
 \city{Xinjiang}
 \country{China}
}

\author{Qiaoming Zhu}
\affiliation{%
  \institution{Soochow University}
  \city{Suzhou}
  \country{China}}

\author{Mang Ye}
\affiliation{%
  \institution{Wuhan University}
  \city{Wuhan}
  \country{China}}
\renewcommand{\shortauthors}{Xinyu Zhou et al.}

\begin{abstract}
Multilingual Text-Based Person Retrieval (TBPR) aims to retrieve individuals with texts in multiple languages. Despite its practical importance, it has been overlooked by existing research, which mainly focuses on English-only retrieval.
The only prior attempt at multilingual TBPR suffers from (1) the absence of a large-scale, multilingual pretraining dataset and (2) a suboptimal model architecture for multilingual text encoding and cross-modal alignment, which employs a shared encoder that overlooks linguistic heterogeneity and relies on English as a pivot for cross-modal alignment, introducing English-centric bias.
For this, we contribute MultiPedes, the first large-scale, pedestrian-centric multilingual pretraining dataset. It contains 10 million image–text pairs across four languages, constructed by a multilingual quality-aware data construction pipeline.
We also propose ViSeLA (Vision-grounded Semantic Learning and Alignment), a novel framework featuring: (1) a semantic-aware multilingual learning module that extends the mixture-of-experts architecture by routing language-invariant semantics to specialized experts for improved multilingual feature learning; and (2) a vision-grounded cross-modal alignment module that employs a semantic probe to capture fine-grained semantics across modalities, enabling direct cross-modal alignment without relying on a pivot language.
Extensive experiments on both multilingual and English-only TBPR datasets demonstrate the effectiveness of ViSeLA. The dataset and code are available at https://github.com/Flame-Chasers/ViSeLA.
\end{abstract}





\maketitle

\begin{figure}[ht]
\centering
\includegraphics[width=0.46\textwidth]{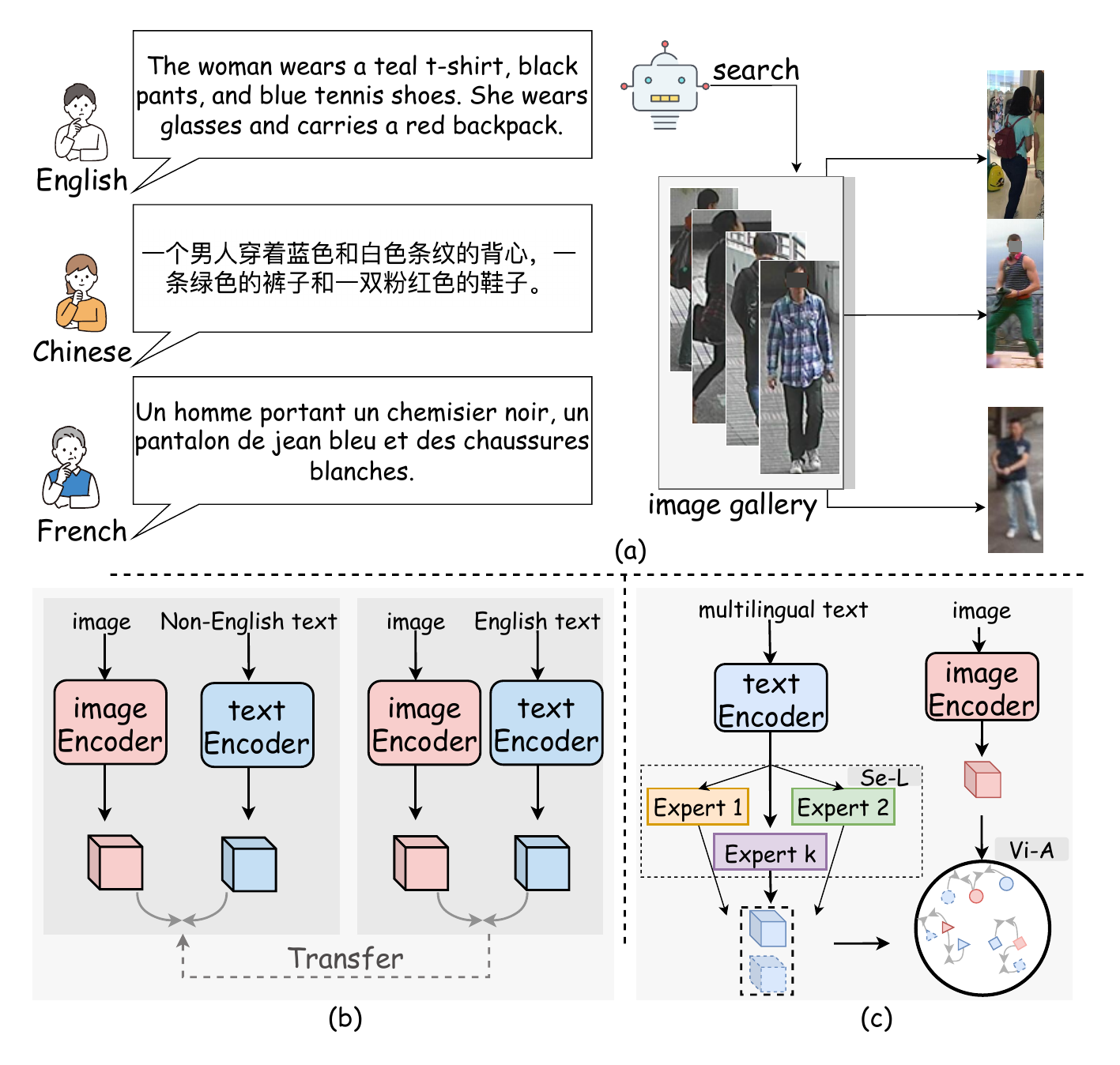}
\caption{
(a) Multilingual TBPR, where person images are retrieved with texts in multiple languages.
(b) Existing method adopts a shared text encoder for all languages and relies on an English-centered knowledge transfer paradigm for cross-lingual cross-modal alignment.
(c) Our ViSeLA framework introduces Se-L module for multilingual learning and Vi-A module for vision-grounded cross-modal alignment.
}
\label{fig:intro}
\end{figure}

\section{Introduction}
Text-based person retrieval (TBPR)~\cite{li2017person} aims to identify target individuals from an image gallery using natural language descriptions as queries. Compared with person re-identification (Re-ID)~\cite{ye2021deep}, which typically relies on image-based queries, TBPR leverages more accessible and flexible textual descriptions, making it more suitable for real-world scenarios—such as intelligent surveillance and public security—where query images are often unavailable, thereby attracting growing research interest~\cite{cao2024empirical, jiang2025modeling}.

The broad applicability of TBPR in real-world scenarios naturally motivates multilingual TBPR~\cite{cao2025multilingual} (Figure~\ref{fig:intro}(a)), which enables retrieval with textual queries in multiple languages—a necessity for serving linguistically diverse users. However, research in this task remains scarce and faces two major challenges.
\ding{182} There is a lack of large-scale, pedestrian-centric pretraining datasets with multilingual textual annotations. While recent works~\cite{tan2024harnessing, jiang2025modeling, cao2025empirical} highlight the effectiveness of pretraining on such datasets in boosting TBPR performance, their annotations are almost exclusively in English, rendering them inadequate for multilingual pretraining.
\ding{183} For multilingual TBPR, effective multilingual text encoding and cross-modal alignment are pivotal. However, the current method, Bi-IRRA~\cite{cao2025multilingual}, exhibits limitations in these regards.
In text encoding, it adopts a shared encoder under a homogeneity assumption of a unified cross-lingual space, neglecting intrinsic linguistic heterogeneity (\emph{e.g.}, divergent syntactic structures) and thus failing to capture language-specific features vital for robust multilingual performance.
In cross-modal alignment, it uses English text–image pairs as a teacher in knowledge distillation to align non-English texts with visual content. This imposes English-centric semantic granularity and structural priors on other languages, introducing an English-centric bias that undermines cross-lingual fairness.

To handle the first challenge, we propose MultiPedes, a large-scale multilingual pedestrian-centric dataset comprising $10$ million image–text pairs across four typologically diverse languages (\emph{i.e.}, English, Chinese, French, and German).
To ensure high-quality cross-modal alignment while scaling to multiple languages at low cost, we design a multilingual quality-aware data construction pipeline. This pipeline first employs an consensus-guided filtering strategy to establish reliable semantic consistency in English-centric data. Building upon this, we perform machine translation—an increasingly robust and cost-effective process enabled by recent advances in large language models—to obtain multilingual descriptions, ensuring strong alignment between images and the translated texts.
As a result, MultiPedes is a high-quality and scalable multilingual dataset, providing a strong foundation for learning robust and generalizable pedestrian features across languages.

In response to the second challenge, we propose a Vision-Grounded Semantic Learning and Alignment (ViSeLA) framework, comprising two components: a Semantic-aware multilingual learning (Se-L) module for self-adaptive multilingual encoding, and a Vision-grounded Cross-modal Alignment (Vi-A) module for fine-grained cross-modal alignment.
Se-L extends the classical Mixture-of-Experts (MoE) architecture with semantic-aware specialization.
It employs a routing mechanism that dynamically assigns textual tokens to experts based on underlying semantics. This ensures that each expert processes semantically coherent yet linguistically diverse inputs, implicitly aligning equivalent concepts across languages. 
Vi-A operates on paired image–text data, where each image is paired with a caption in an arbitrary language. The image thus serves as a semantic anchor. Based on it, Vi-A introduces a semantic probe to capture fine-grained cues from both visual and textual modalities. By directly aligning multilingual textual features to the shared visual anchor, Vi-A implicitly enforces cross-modal alignment centered on image semantics, effectively mitigating English-centric bias.

Our key contributions are:
(1) We introduce MultiPedes, the first large-scale pedestrian-centric multilingual pretraining dataset, containing $10$ million image–text pairs across four languages, constructed via a multilingual quality-aware data construction pipeline to support scalable multilingual TBPR.
(2) We propose ViSeLA, a framework that addresses linguistic heterogeneity and English-centric bias through two key components: a Se-L module for self-adaptive multilingual encoding, and a Vi-A module for fine-grained cross-modal alignment.
(3) Extensive experiments on multiple multilingual TBPR benchmarks validate the effectiveness and generality of ViSeLA, with consistent state-of-the-art performance.

\section{Related Work}
\subsection{Text-based Person Retrieval}
Text-based person retrieval (TBPR)~\cite{li2017person} aims to retrieve pedestrian images from natural language descriptions. Early methods~\cite{farooq2020convolutional, zheng2020dual} extract global visual and textual features using separate backbones and align them in a shared space, but often overlook fine-grained attributes critical for discrimination. Later works~\cite{jing2020pose, wang2020vitaa} introduce local-matching with external supervision (\emph{e.g.}, human parsing) to align fine-grained cross-modal features, improving retrieval performance. With the advent of Vision-Language Pretraining (VLP) models~\cite{radford2021learning}, their strong cross-modal alignment largely supersedes earlier models, prompting research on effectively leveraging VLP capabilities~\cite{cao2024empirical}.
Recently, the domain gap between generic corpora used in VLP pretraining and pedestrian-centric data for TBPR has drawn increasing attention. In response, a growing body of work~\cite{yang2023towards, tan2024harnessing} has focused on constructing large-scale, person-specific pretraining datasets to enhance both retrieval performance and generalization.

\begin{figure*}[t]
\centering
\includegraphics[width=\textwidth]{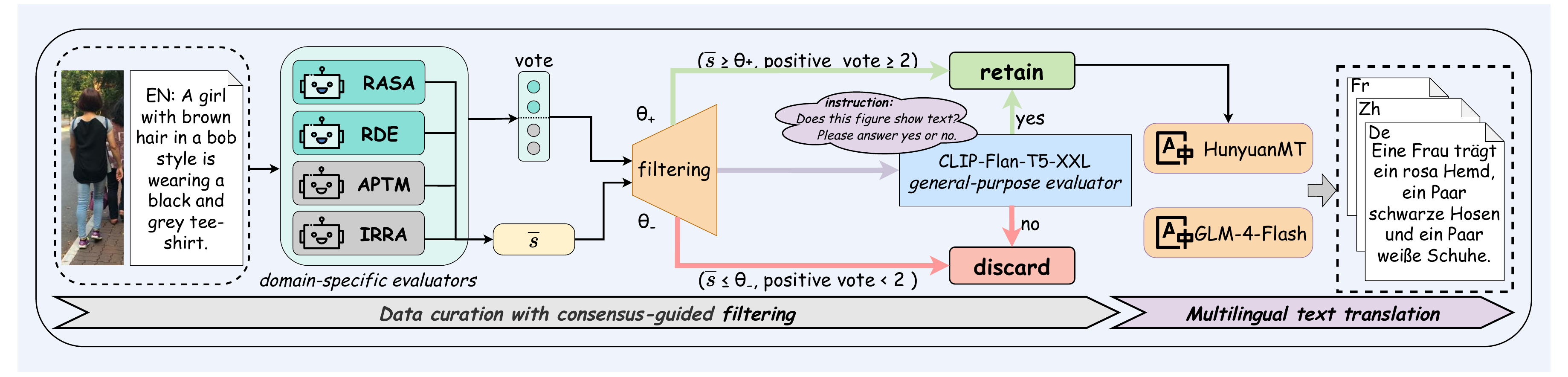}
\caption{Multilingual quality-aware data construction pipeline consisting of two stages: (1) data curation with consensus-guided filtering, and (2) multilingual text translation. In the first stage, consensus-guided filtering leverages both domain-specific and domain-agnostic evaluators to identify and remove noisy samples from large-scale image–text pairs collected from public datasets, resulting in higher-quality data. In the second stage, large language models are employed to perform machine translation at scale, generating multilingual descriptions and constructing the MultiPedes dataset.}
\label{fig:data}
\end{figure*}

Moreover, multilingual TBPR has recently emerged as a practical research direction. Bi-IRRA~\cite{cao2025multilingual}, the first work in this setting, transfers alignment knowledge from English--image pairs to non-English data through cross-lingual distillation, but remains English-centric bias and does not address scalability. In contrast, we introduce the scalable MultiPedes pretraining dataset and the ViSeFA framework to jointly improve multilingual representation learning and cross-modal alignment.

\subsection{Multilingual Image-Text Retrieval}
Multilingual Image--Text Retrieval (MITR)~\cite{li2023unifying} extends cross-modal retrieval by enabling visual retrieval with queries in multiple languages. A major challenge is the scarcity of large-scale multilingual image--text data. Existing approaches generally follow two directions: some methods~\cite{ni2021m3p, jain2021mural} combine heterogeneous data, including English image--text pairs and multilingual parallel text corpora, while others~\cite{zhou2021uc2} use machine translation (MT) to generate non-English captions from existing English image--text pairs. These data sources lead to different alignment strategies: heterogeneous-data methods typically use English text as an anchor to bridge languages and modalities, whereas MT-based methods often use images as a shared anchor for cross-lingual and cross-modal alignment.

Unlike general-purpose MITR, multilingual TBPR requires more fine-grained semantic alignment to distinguish visually similar pedestrians. Accordingly, we propose ViSeFA, which employs fine-grained visual semantics as the alignment anchor to jointly facilitate multilingual feature learning and precise cross-modal alignment.

\section{Dataset}
In this section, we introduce the proposed multilingual pedestrian-centric pretraining dataset \textbf{MultiPedes}, including its construction pipeline and key statistics.

\begin{table}[tbh]
    \centering
    \caption{Comparison with other mainstream pretraining datasets for TBPR. \# denotes the number of samples, and Avg. Len indicates the average text length.}
    \label{tab:DatasetComparsion}
    \resizebox{\linewidth}{!}{%
    \renewcommand\arraystretch{1.2}
    \setlength{\extrarowheight}{3pt}

        \begin{tabular}{l|c|c|c|c|c|c}
        \hline\thickhline
        \rowcolor{lightgray}
        \textbf{Dataset} & \textbf{\#Images} & \textbf{\#Texts} & \textbf{Avg. Len} & \textbf{Language} & \textbf{TTR} & \textbf{CLIPScore} \\
        \hline\hline
        LUPerson-T~\cite{shao2023unified} & 1.30M & 1.30M & 27.39 & En & 0.88 & 0.22 \\
        LUPerson-M~\cite{tan2024harnessing} & 1.00M & 4.00M & 28.61 & En & 0.80 & 0.36 \\
        HAM-PEDES~\cite{jiang2025modeling} & 1.00M & 2.00M & 26.30 & En & 0.80 & 0.40 \\
        SynTBPR~\cite{cao2025empirical} & 1.63M & 6.62M & 34.10 & En & 0.81 & 0.43 \\
        WebPerson~\cite{zheng2025gradient} & 5M & 10M & 23.34 & En & 0.81 & 0.40 \\
        \hline
        MultiPedes & 1.25M & 10.00M & 38.11 & \makecell[c]{En, ZH, \\ Fr, De} & 0.87 & 0.40 \\
        \hline\thickhline
        \end{tabular}
    }
\end{table}
\subsection{Construction Pipeline}
To efficiently construct high-quality multilingual textual annotations at scale, we design a multilingual quality-aware data construction pipeline built upon existing English-centric TBPR data. Leveraging recent advances in machine translation, the pipeline enables scalable and cost-effective multilingual data generation while preserving semantic fidelity.
The pipeline comprises two stages: (1) data curation with consensus-guided filtering, and (2) multilingual text translation as shown in Figure~\ref{fig:data}.

\textbf{Data Curation with Consensus-Guided Filtering.}
We first collect English image–text pairs from publicly available pedestrian-centric pretraining datasets~\cite{jiang2025modeling, zheng2025gradient} as the basis for multilingual expansion. To ensure high semantic consistency between images and the translated multilingual texts, we introduce a consensus-guided filtering strategy to remove potentially noisy pairs.

For each collected image–text pair \((I, T^{\mathrm{en}})\), multiple domain-specific evaluators independently assess semantic consistency and assign similarity scores \(s_i\). The evaluators are executed in parallel, incurring minimal additional runtime, while the extra computation devoted to noise filtering leads to higher-quality pretraining data, constituting a favorable quality–efficiency trade-off, as further evidenced by the supplementary material.
Then, given the predefined thresholds $\theta_{+}$ and $\theta_{-}$, the $i$-th evaluator casts a positive vote if $s_i > \theta_{+}$ and a negative vote if $s_i < \theta_{-}$. Let $\bar{s}$ denote the mean similarity score across evaluators. A sample pair is retained if it receives at least two positive votes and $\bar{s} > \theta_{+}$, and is discarded if it receives at least two negative votes and $\bar{s} < \theta_{-}$. 
Through consensus among domain-specific evaluators, the majority of samples can be reliably determined, while the remaining ambiguous cases are forwarded for secondary verification by a domain-agnostic evaluator. We implement this evaluator using a VQA model~\cite{lin2024evaluating} to make the final decision, prompting it with: 
\textit{“Does this figure <img> show ‘<text>’? Please answer yes or no.”}, 
where the placeholder <img> and <text> correspond to the image $I$ and its description $T^{\mathrm{en}}$, respectively. Samples receiving a “yes” response are retained, while those receiving a “no” response are discarded. 
Ultimately, we retain $2.5$M high-quality image–text pairs for subsequent dataset construction.

Overall, the consensus-guided filtering prioritizes domain-specific evaluators as the primary filter applied to all samples, where their rich domain knowledge and consensus mechanism enable reliable and fine-grained semantic assessment. A domain-agnostic evaluator is then invoked for a small subset of ambiguous cases to provide complementary judgment, thereby fully exploiting domain knowledge for reliable decisions while mitigating potential biases introduced by domain-specific modeling.

\textbf{Multilingual Text Translation.}
Due to the high semantic consistency of the curated English image–text pairs, we can reliably extend the dataset to multiple languages via a simple yet effective translation procedure. We select three widely used languages beyond English—Chinese, French, and German—as target languages.

To enhance linguistic diversity and avoid homogenized expressions, we employ multiple translation models~\cite{5team2025glm45agenticreasoningcoding, hunyuan_mt} and randomly select one for each translation instance. Concretely, for each English description $T^{\mathrm{en}}$, we prompt the translation model with the following instruction:
\textit{Translate the following English person description into \texttt{<lang\_name>}.  
Requirements:• Faithfully translate all details about clothing, colors, accessories, etc.  
• Preserve the original descriptive order and logical structure.  
• Use natural and fluent \texttt{<lang\_name>} expressions.  
}
Here, \texttt{<lang\_name>} denotes the target language name (\emph{e.g.}, Chinese, French, German). Given the strong semantic consistency of the translated texts (shown in Sec.~\ref{sec: statistics}), we do not apply additional post-translation filtering.

As a result, each English description is expanded into corresponding multilingual versions in all three target languages, leading to the construction of MultiPedes, a large-scale multilingual dataset comprising $10$ million image–text pairs. We provide illustrative examples in the supplementary material.

\subsection{Statistics}
\label{sec: statistics}
\begin{figure}[t]
\centering
\includegraphics[width=0.42\textwidth]{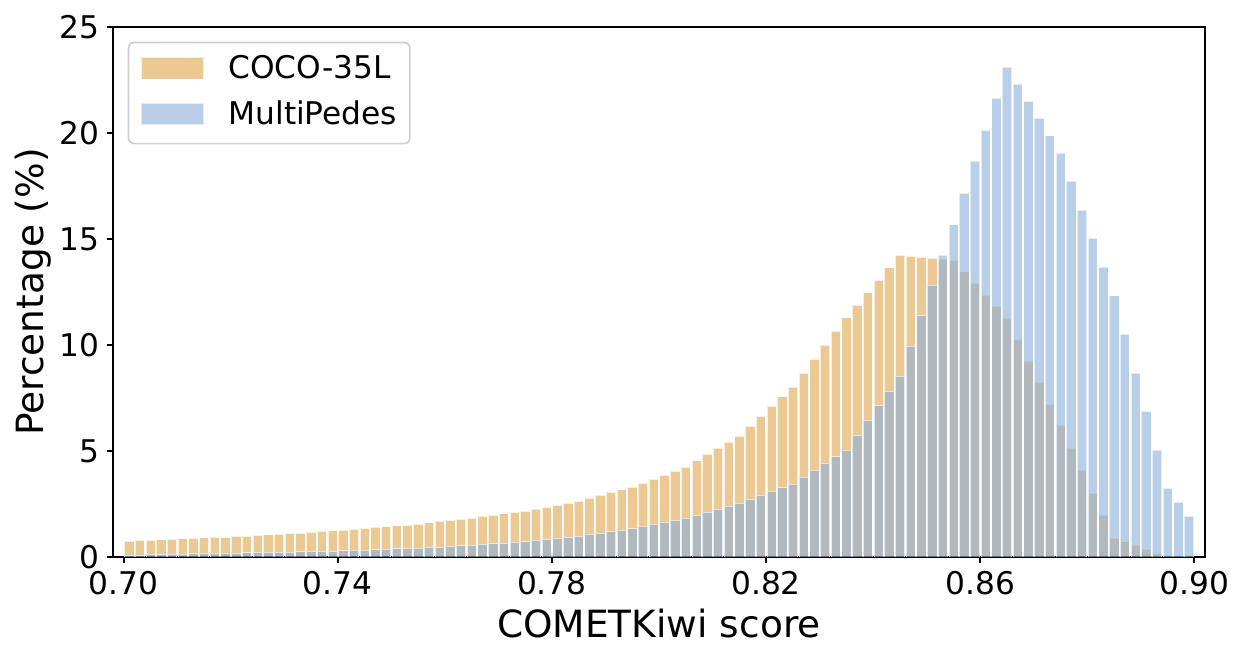}
\caption{Comparison of COMET-Wiki score distributions between MultiPedes and COCO-35L~\cite{thapliyal2022crossmodal}. The scores are averaged over three target languages for each sample.}
\label{fig:comet}
\end{figure}
We present the statistics of the MultiPedes dataset from two perspectives: \textbf{dataset properties} and \textbf{data quality}, and compare it with other mainstream pedestrian-centric pretraining datasets. 

In terms of \textbf{dataset properties}, MultiPedes exhibits two prominent characteristics: large scale and multilingual coverage. 
MultiPedes contains \(10\) million image–text pairs, making it one of the largest pedestrian-centric pretraining dataset for TBPR. 
Moreover, unlike prior datasets limited to English-only annotations, MultiPedes is the first to provide large-scale multilingual descriptions spanning English, Chinese, French, and German, facilitating effective and scalable pretraining for multilingual TBPR.

In terms of \textbf{data quality}, we analyze MultiPedes from three aspects: textual diversity, image–text alignment, and cross-lingual consistency. 
Textual diversity is measured using the Type–Token Ratio (TTR), defined as the ratio of unique tokens to the total number of tokens, which reflects the richness of vocabulary usage. 
Image–text alignment is evaluated on the image–text pairs of MultiPedes with CLIPScore~\cite{radford2021learning}, measuring the semantic consistency between images and their corresponding English descriptions. As shown in Table~\ref{tab:DatasetComparsion}, MultiPedes achieves decent performance in both TTR and CLIPScore, indicating good textual diversity and solid image–text semantic alignment.
For cross-lingual consistency, we compute COMET-Wiki~\cite{rei2023scaling} scores between English and its translations in Chinese, French, and German. COMET-Wiki is a reference-free metric for assessing semantic equivalence across languages. As shown in Figure~\ref{fig:comet}, we compare the score distributions of MultiPedes with those of COCO-35L~\cite{thapliyal2022crossmodal}, a widely used public dataset with parallel multilingual text descriptions, where scores are averaged across languages. The results indicate that MultiPedes exhibits stronger cross-lingual consistency.

\begin{figure*}[ht]
\centering
\includegraphics[width=\textwidth]{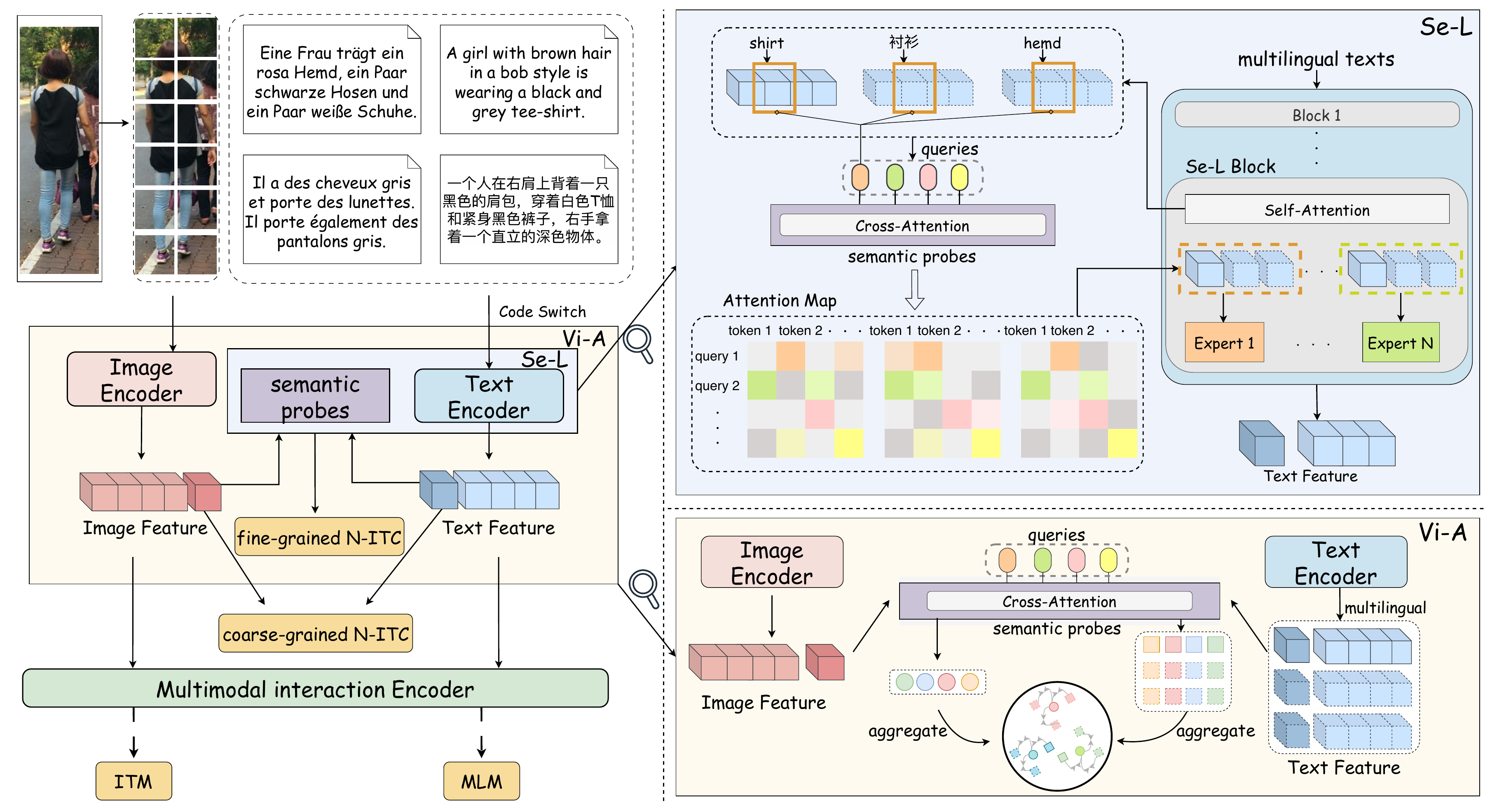}
\caption{Overview of our proposed ViSeLA framework. It consists of two key submodules: (1) the Se-L module, which leverages semantic information to route input text features to specialized experts, thereby extracting language-invariant textual features; and (2) the Vi-A module, which employs the semantic probe to capture fine-grained cross-modal semantics and enables image-grounded cross-modal alignment.}
\label{fig:method}
\end{figure*}

\section{Method}
In this section, we elaborate the proposed ViSeLA, which comprises a Se-L module for improved multilingual learning and a Vi-A module for fine-grained cross-modal alignment, as illustrated in Figure~\ref{fig:method}.

\subsection{Baseline}
We adopt X\(^2\)-VLM~\cite{zeng2023x} as our baseline model, which comprises an image encoder, a text encoder, and a multimodal interaction encoder. Both the image and text encoders are implemented as \(12\)-layer Transformer, responsible for extracting unimodal visual and textual features, respectively. The multimodal interaction encoder consists of \(6\)-layer Transformer and leverages cross-attention mechanisms to model cross-modal interactions.

The model takes image–text pairs \((I, T^{\ell})\) as input, where \(T^{\ell}\) denotes a caption written in language \(\ell\), where \(\ell \in \mathcal{A} = \{\text{en}, \text{zh}, \text{fr}, \text{de}\}\), corresponding to English, Chinese, French, and German, respectively. 
Given an image \(I\), it is first partitioned into \(N_p\) non-overlapping patches and encoded by a vision encoder to produce a sequence of visual features \( F_v = \{v_{\text{cls}}, v_1, \ldots, v_{N_p}\}. \) The textual input \(T^{\ell}\) is augmented with a code-switching strategy~\cite{ni2021m3p}, in which words are randomly replaced with their counterparts from other languages with a certain probability. This augmentation introduces cross-lingual lexical variations and encourages the model to learn language-invariant semantic features, thereby improving multilingual generalization. The augmented text is then processed by the text encoder to yield textual features \(F_t = \{t_{\text{cls}}, t_1, \ldots, t_L\}.\)
The global image feature \(v_{\text{cls}}\) and the global text feature \(t_{\text{cls}}\) are subsequently used to compute a global image–text similarity score \(S^{c}\) via cosine similarity. Based on this score, we optimize a normalized image–text contrastive loss (N-ITC)~\cite{cao2024empirical}, denoted as \(\mathcal{L}^{c}_{N-ITC}\). The unimodal features \(F_v\) and \(F_t\) are fed into a multimodal interaction encoder to facilitate cross-modal learning, which is optimized jointly via the image–text matching loss (ITM), denoted as \(\mathcal{L}_{\text{ITM}}\) and the masked language modeling loss (MLM), denoted as \(\mathcal{L}_{\text{MLM}}\).

\subsection{Semantic-aware multilingual learning}
To enhance multilingual modeling capacity and promote language-invariant textual features, we propose a Semantic-aware multilingual Learning (Se-L) module. Se-L integrates a semantic-aware MoE into selected ViT blocks of the text encoder, referred to as Se-L blocks. It consists of \(N\) semantic-specialized experts \(\mathcal{E} = \{\epsilon_j \mid j \in [1, N]\}\) and the semantic probe from the Vi-A module (detailed in Section~\ref{sec:Vi-A}), which are composed of \(N\) learnable queries and a cross-attention layer, serving as a semantic-aware router that dynamically selects and combines experts based on textual semantics.

Specifically, given the textual features \(F_t = \{t_1, \dots, t_L\}\) output by the self-attention layer within an Se-L block, the semantic-aware router computes a cross-attention map \(P \in \mathbb{R}^{L \times N}\), where each entry \(P_{i,j}\) denotes the attention weight of the \(j\)-th query of the semantic probe on the \(i\)-th token. Since each probe corresponds to a distinct semantic concept, \(P\) effectively represents the relevance between \(t_i\) and each semantic cue.

\begin{table*}[h]\small
\centering
\caption{Performance comparison with state-of-the-art methods on multilingual TBPR datasets. 
We categorize these methods into two groups based on whether they are pre-trained on large-scale person data (per their original papers): Bi-IRRA uses an English-only dataset (LUPerosn-M~\cite{tan2024harnessing}), while ViSeLA leverages our proposed MultiPedes dataset.
Results are reported for four languages and their average. The best results are highlighted in bold, and the second-best results are underlined.}
\label{tab:compact_comparison}
\small{
\resizebox{\linewidth}{!}{%
\renewcommand\arraystretch{1.2}
\begin{tabular}{l|l|ccc|ccc|ccc|ccc|ccc}
\hline\thickhline
\rowcolor{lightgray}
 &  & \multicolumn{3}{c|}{\textbf{English}} & \multicolumn{3}{c|}{\textbf{Chinese}} & \multicolumn{3}{c|}{\textbf{French}} & \multicolumn{3}{c|}{\textbf{German}} & \multicolumn{3}{c}{\textbf{Avg}} \\
\rowcolor{lightgray}
\multirow{-2}{*}{\textbf{Dataset}} & \multirow{-2}{*}{\textbf{Method}} & R1 & R10 & mAP & R1 & R10 & mAP & R1 & R10 & mAP & R1 & R10 & mAP & R1 & R10 & mAP \\
\hline\hline

 & \multicolumn{16}{l}{\textit{Without Pre-training on Large-scale Person Data:}} \\
\cline{2-17}
\multirow{7}{*}{CUHK-PEDE(M)}
 & CCLM~\cite{zeng2022cross} & 74.68 & 93.42 & 65.78 & 70.83 & 92.85 & 63.14 & 71.17 & 92.30 & 62.89 & 70.19 & 92.24 & 62.47 & 71.72 & 92.70 & 63.57 \\
 & X²-VLM~\cite{zeng2023x} & 76.97 & 94.96 & 67.86 & 74.81 & 94.17 & 66.00 & 75.18 & 94.18 & 66.17 & 74.71 & 94.30 & 65.73 & 75.42 & 94.40 & 66.44 \\
 & CCRK~\cite{nie2024improving} & 70.11 & 92.53 & 62.10 & 68.31 & 91.91 & 60.69 & 68.29 & 91.36 & 60.33 & 67.01 & 90.74 & 59.54 & 68.43 & 91.64 & 60.67 \\
 & Bi-IRRA~\cite{cao2025multilingual}  & \textbf{78.82} & \textbf{95.47} & \underline{69.68} &\textbf{ 76.43} & \textbf{94.79} & \underline{67.79} & \underline{76.46} & \textbf{94.35} & \underline{67.26} & \underline{75.57} & \textbf{94.31} & \underline{67.07} & \underline{76.82} & \underline{94.73} & \underline{67.95} \\
\rowcolor[HTML]{D7F6FF}
\cellcolor{white} & ViSeLA & \underline{78.65} & \underline{95.21} & \textbf{70.33} & \underline{76.08} & \underline{94.17} & \textbf{68.30} & \textbf{76.65} & \underline{94.23} & \textbf{68.35} & \textbf{75.92} & \underline{94.18} & \textbf{67.87} & \textbf{76.83} & \textbf{94.45} & \textbf{68.71} \\
\cline{2-17}
 & \multicolumn{16}{l}{\textit{With Pre-training on Large-scale Person Data:}} \\
\cline{2-17}
\multirow{3}{*}{}
 & Bi-IRRA~\cite{cao2025multilingual} & \textbf{79.43} & \textbf{95.68} & \underline{70.51} & \underline{76.55} & \textbf{94.49} & \underline{68.05} & \underline{76.60} & \textbf{94.31} & \underline{67.65} & \underline{75.79} & \textbf{94.38} & \underline{67.89} & \underline{77.09} & \textbf{94.72} & \underline{68.53} \\
\rowcolor[HTML]{D7F6FF}
\cellcolor{white} & ViSeLA & \underline{79.37} & \underline{95.11} & \textbf{71.68} & \textbf{77.29} & \underline{93.94} & \textbf{69.72} & \textbf{77.66} & \underline{94.10} & \textbf{69.71} & \textbf{77.45} & \underline{94.02} & \textbf{69.53} & \textbf{77.94} & \underline{94.29} & \textbf{70.16}  \\
\hline\hline

 & \multicolumn{16}{l}{\textit{Without Pre-training on Large-scale Person Data:}} \\
\cline{2-17}
\multirow{7}{*}{ICFG-PEDES(M)}
 & CCLM~\cite{zeng2022cross} & 64.14 & 85.50 & 35.53 & 61.83 & 83.98 & 34.01 & 60.95 & 83.58 & 33.16 & 61.42 & 83.72 & 33.84 & 62.09 & 84.20 & 34.14 \\
 & X²-VLM~\cite{zeng2023x} & 68.39 & 87.58 & 40.57 & 66.54 & 86.28 & 39.13 & 65.74 & 85.69 & 38.59 & 66.34 & 86.02 & 38.77 & 66.75 & 86.39 & 39.27 \\
 & CCRK~\cite{nie2024improving} & 60.31 & 83.37 & 32.32 & 57.98 & 81.99 & 30.91 & 57.57 & 81.65 & 30.20 & 56.52 & 80.63 & 29.28 & 58.10 & 81.91 & 30.68 \\
 & Bi-IRRA~\cite{cao2025multilingual}  & \underline{68.53} & \underline{87.79} & \underline{41.82} & \underline{66.79} & \underline{86.44} & \underline{40.26} & \underline{66.90} & \underline{86.40} & \underline{39.44} & \underline{67.05} & \underline{86.62} & \underline{39.91} & \underline{67.32} & \underline{86.81} & \underline{40.36} \\
\rowcolor[HTML]{D7F6FF}
\cellcolor{white} & ViSeLA & \textbf{70.28} & \textbf{87.80} & \textbf{43.57} & \textbf{68.21} & \textbf{86.49} & \textbf{41.97} & \textbf{67.60} & \textbf{86.47} & \textbf{41.45} & \textbf{68.11} & \textbf{86.71} & \textbf{41.79} & \textbf{68.55} & \textbf{86.87} & \textbf{42.20} \\
\cline{2-17}
 & \multicolumn{16}{l}{\textit{With Pre-training on Large-scale Person Data:}} \\
\cline{2-17}
\multirow{3}{*}{}
 & Bi-IRRA~\cite{cao2025multilingual}  & \underline{70.63} & \underline{88.47} & \underline{43.15} & \underline{67.81} & \underline{86.61} & \underline{41.24} & \underline{67.54} & \underline{86.86} & \underline{41.00} & \underline{67.51} & \underline{86.86} & \underline{41.09} & \underline{68.37} & \underline{87.20} & \underline{41.62} \\
\rowcolor[HTML]{D7F6FF}
\cellcolor{white} & ViSeLA & \textbf{71.35} & \textbf{88.51} & \textbf{44.83} & \textbf{69.26} & \textbf{87.03} & \textbf{43.37} & \textbf{68.80} & \textbf{87.16} & \textbf{42.55} & \textbf{69.23} & \textbf{87.10} & \textbf{43.02} & \textbf{69.66} & \textbf{87.45} & \textbf{43.45} \\
\hline\hline

 & \multicolumn{16}{l}{\textit{Without Pre-training on Large-scale Person Data:}} \\
\cline{2-17}
\multirow{7}{*}{UFineBench(M)}
 & CCLM~\cite{zeng2022cross} & 85.86 & 96.62 & 84.19 & 84.98 & 96.44 & 83.37 & 83.58 & 95.84 & 81.80 & 83.42 & 95.77 & 81.49 & 84.46 & 96.17 & 82.71 \\
 & X²-VLM~\cite{zeng2023x} & 89.32 & 98.02 & 87.82 & 88.20 & \underline{97.90} & 87.12 & 87.98 & 97.39 & 86.21 & 86.86 & 97.29 & 85.52 & 88.09 & 97.65 & 86.67 \\
 & CCRK~\cite{nie2024improving} & 81.06 & 95.14 & 79.21 & 79.58 & 94.49 & 77.98 & 78.61 & 94.20 & 76.84 & 77.63 & 93.98 & 75.87 & 79.22 & 94.45 & 77.48 \\
 & Bi-IRRA~\cite{cao2025multilingual} & \underline{90.45} & \underline{98.18} & \underline{89.66} & \underline{89.98} & \textbf{98.05} & \underline{89.22} & \underline{89.30} & \underline{97.73} & \underline{88.32} & \underline{89.39} & \textbf{97.75} & \underline{88.02} & \underline{89.78} & \underline{97.93} & \underline{88.81} \\
\rowcolor[HTML]{D7F6FF}
\cellcolor{white} & ViSeLA & \textbf{91.45} & \textbf{98.19} & \textbf{90.18} & \textbf{90.41} & \textbf{98.05} & \textbf{89.34} & \textbf{89.97} & \textbf{97.89} & \textbf{88.87} & \textbf{89.87} & \underline{97.66} & \textbf{88.61} & \textbf{90.42} & \textbf{97.95} & \textbf{89.25} \\
\cline{2-17}
 & \multicolumn{16}{l}{\textit{With Pre-training on Large-scale Person Data:}} \\
\cline{2-17}
\multirow{2}{*}{}
 & Bi-IRRA~\cite{cao2025multilingual}  & \underline{91.26} & \underline{98.26} & \underline{90.01} & \underline{90.67} & \underline{98.13} & \underline{89.78} & \underline{89.93} & \underline{97.88} & \underline{89.03} & \underline{89.34} & \underline{97.92} & \underline{88.32} & \underline{90.30} & \underline{98.05} & \underline{89.29} \\
\rowcolor[HTML]{D7F6FF}
\cellcolor{white} & ViSeLA & \textbf{92.27} & \textbf{98.35} & \textbf{91.34} & \textbf{91.55} & \textbf{98.20} & \textbf{90.60} & \textbf{91.17} & \textbf{98.09} & \textbf{90.20} & \textbf{91.10} & \textbf{97.99} & \textbf{90.08} & \textbf{91.52} & \textbf{98.16} & \textbf{90.56} \\

\hline\thickhline
\end{tabular}
}}
\label{tab:multilingual tbpr}
\end{table*}

For each texutal token \(t_i\), we route it to its most relevant experts based on \(P\). First, we sort the attention scores \(\{P_{i,j}\}_{j=1}^{N}\) between token \(t_i\) and all queries in descending order, and select the smallest index set \(\mathcal{K}_i \subseteq \{1,\dots,N\}\) such that:
\begin{equation} 
\sum_{k \in \mathcal{K}i} P_{i,k} \geq \delta,
\end{equation}
where \(\delta \in (0,1]\) is a pre-defined threshold that controls the sparsity and semantic coverage of expert selection. Subsequently, we compute normalized routing weights over the selected experts as:
\begin{equation} 
w_{i,j} = \begin{cases} \frac{\exp(P_{i,j})}{\sum_{k \in \mathcal{K}_i} \exp(P_{i,k})} & \text{if } j \in \mathcal{K}_i \\ 0 & \text{otherwise} \end{cases}
\end{equation}
The output feature for token \(t_i\) is then obtained as a weighted combination of the activated experts:
\begin{equation} 
t_i = \sum_{j} w_{i,j} \cdot \epsilon_j(t_i). 
\end{equation}

During encoding, the same semantic probe captures semantically similar components across texts \(T^l\) in different languages and routes them to the corresponding expert for processing. This design pulls features of shared semantics closer in the feature space, thereby facilitating the learning of language-invariant textual features.
\subsection{Vision-grounded Cross-modal Alignment}
\label{sec:Vi-A}
The Vision-grounded Semantic Alignment (Vi-SA) is a lightweight module for visual-grounded fine-grained cross-modal alignment.

The core of this module is the semantic probe, which consists of \(N\) learnable queries \(\mathcal{Q} = \{q_1, q_2, \ldots, q_N\}\) (matching the number of experts in the Se-L module) and a multi-head cross-attention (MCA) layer, enabling each query to extract the visual or textual features it attends to via cross-attention.
Specifically, given visual features \(F_v\), the cross-attention is performed as follows: 
\begin{equation}
\hat{F}_v = \mathrm{MCA}(\mathcal{Q}, F_v, F_v),
\end{equation}
where \(\hat{F}_v = \{\hat{v}_1, \hat{v}_2, \ldots, \hat{v}_N\}\). Each \(\hat{v}_i\) captures a distinct visual semantic concept (\emph{e.g.} object parts, attributes).
In parallel, the same mechanism is applied to the textual features, yielding \(\hat{F}_t = \{\hat{t}_1, \hat{t}_2, \ldots, \hat{t}_N\}\). 

Visual and textual semantics attended by the same query are regarded as corresponding pairs.
Then, we compute the cosine similarity for each semantic pair and aggregate them as follows:
\begin{equation}
S^f = \log \sum_{i=1}^{N} \exp \left(\frac{\hat{v}_i^\top \hat{t}_i}{\lVert \hat{v}_i \rVert \cdot \lVert \hat{t}_i \rVert}\right).
\end{equation}
Finally, the fine-grained similarity score \(S^f\) is used to optimize a normalized image–text contrastive loss, denoted as \(\mathcal{L}^{f}_{\text{N-ITC}}\), thereby enabling effective fine-grained cross-modal alignment.

\begin{table*}[t]\small
    \centering
    \caption{Performance comparison on English TBPR with state-of-the-art methods on CUHK-PEDES, ICFG-PEDES, and UFineBench. Methods are grouped by whether they use large-scale person-centric pretraining. The best results are highlighted in bold, and the second-best results are underlined.}
    \resizebox{0.88\linewidth}{!}{
        \renewcommand\arraystretch{1.2}
        \begin{tabular}{l|cccc|cccc|cccc}
        \hline\thickhline
        \rowcolor{lightgray}
        & \multicolumn{4}{c|}{CUHK-PEDES} & \multicolumn{4}{c|}{ICFG-PEDES} & \multicolumn{4}{c}{UFineBench} \\
        \rowcolor{lightgray}
        \multirow{-2}{*}{Methods} 
        & R@1   & R@5   & R@10  & mAP   
        & R@1   & R@5   & R@10  & mAP   
        & R@1   & R@5   & R@10  & mAP   \\
        \hline\hline
        \multicolumn{13}{l}{\textit{Without large-scale person-centric pretraining:}} \\
        \hline\hline
        IRRA~\cite{jiang2023cross}        & 73.38 & 89.93 & 93.71 & 66.13  & 63.46 & 80.25 & 85.82 & 38.06  & 83.53 & 92.94 & 95.95 & 82.79 \\
        RaSa~\cite{bai2023rasa}           & \underline{76.51} & 90.29 & 94.25 & \underline{69.38}  & 65.28 & 80.40 & 85.12 & \underline{41.29}  & 86.13 & 94.55 & 96.73 & 84.17 \\
        TBPS-CLIP~\cite{cao2024empirical} & 73.54 & 88.19 & 92.35 & 65.38  & 65.05 & 80.34 & 85.47 & 39.83  & 76.21 & 88.92 & 93.09 & 75.07 \\
        CFAM~\cite{zuo2024ufinebench}     & 75.60 & \underline{90.53} & \underline{94.36} & 67.27  & 65.38 & 81.17 & 86.35 & 39.42  & \underline{88.51} & \underline{95.58} & \underline{97.49} & \underline{87.09} \\
        RDE~\cite{qin2024noisy}           & 75.94 & 90.14 & 94.12 & 67.56  & \underline{67.68} & \textbf{82.47} & \textbf{87.36} & 40.06  & 85.31 & 94.15 & 96.72 & 84.05 \\
        TAG-CLIP~\cite{zhou2026text}       & 74.38 & 88.30 & 92.59 & 67.18  & 66.78 & 81.36 & 86.23 & 43.60 & - & - & - & - \\
        \hline\rowcolor[HTML]{D7F6FF}
        ViSeLA                              & \textbf{77.62} & \textbf{91.26} & \textbf{95.00} & \textbf{69.46}  
                                          & \textbf{67.94} & \underline{81.99} & \underline{86.79} & \textbf{41.72}  
                                          & \textbf{90.10} & \textbf{96.34} & \textbf{97.85} & \textbf{88.76} \\
        \hline\hline
        \multicolumn{13}{l}{\textit{With large-scale person-centric pretraining:}} \\
        \hline\hline
        PLIP~\cite{zuo2023plip}           & 75.36 & 90.86 & 94.87 & -      & 66.17 & 83.37 & \textbf{88.94} & -      & 78.89 & 91.02 & 94.65 & 76.89 \\
        APTM~\cite{yang2023towards}       & 76.53 & 90.04 & 94.15 & 66.91  & 68.51 & 82.99 & 87.56 & 41.22  & 87.59 & 95.42 & 97.22 & 85.75 \\
        AUL~\cite{li2024adaptive}         & 77.23 & 90.43 & 94.41 & -      & 69.16 & 83.32 & 88.37 & -      & 84.60 & 94.28 & 96.69 & 83.06 \\
        MLLM+IRRA~\cite{tan2024harnessing}& 76.82 & 91.16 & 94.46 & 69.55  & 67.05 & 82.16 & 87.33 & 41.51  & 89.13 & 96.10 & 97.90 & 88.38 \\
        HAM~\cite{jiang2025modeling}  & \underline{77.99} & 91.34 & 95.03 & \underline{69.72} & \underline{69.95} & \textbf{83.88} & \underline{88.39} & \underline{42.72}  & \underline{90.94} & \underline{96.77} & \underline{98.24} & \underline{90.13} \\
        CAMeL~\cite{yu2025camel}                & 77.24 & \underline{91.80} & \underline{95.16} & 68.32  & 68.70 & 83.11 & 88.32 & 41.58  & 78.86 & 92.79 & 96.17 & 78.87 \\
        MRA~\cite{yang2026minimizing} & 77.21 & 90.66 & 94.46 & 68.50 & 68.93 & 83.46 & 88.07 & 41.38 & - & - & - & - \\
        \hline\rowcolor[HTML]{D7F6FF}
        ViSeLA                              & \textbf{79.42} & \textbf{92.11} & \textbf{95.48} & \textbf{71.53}  
                                          & \textbf{70.53} & \underline{83.52} & 88.10 & \textbf{44.65}  
                                          & \textbf{91.65} & \textbf{97.00} & \textbf{98.26} & \textbf{90.83} \\
        \hline\thickhline
        \end{tabular}
    }
    \label{tab:english_tbpr}
\end{table*}

During alignment, each image \(I\) is associated with multilingual captions \(T^{\ell}\), allowing all language-specific texts to align directly to the same visual instance. In this process, the image acts as a semantic anchor, enabling the semantic probe to capture fine-grained cues across modalities and effectively enforcing vision-grounded cross-modal alignment, which mitigates English-centric bias.

\subsection{Optimization Objectives}
During training, we jointly optimize multiple objectives to encourage both cross-modal alignment and robust textual feature learning:
\begin{equation}
\mathcal{L} = \mathcal{L}^{c}_{\text{N-ITC}} + \mathcal{L}^{f}_{\text{N-ITC}} + \mathcal{L}_{\text{MLM}} + \lambda \mathcal{L}_{\text{ITM}},
\end{equation}
where \(\lambda\) is a hyper-parameter.

\section{Experiments}
\subsection{Datasets}
We conduct experiments on both traditional English-only TBPR and multilingual TBPR tasks.
For English-only TBPR, the evaluations are performed on three widely used benchmarks, including CUHK-PEDES~\cite{li2017person}, ICFG-PEDES~\cite{ding2021semantically}, and UFineBench~\cite{zuo2024ufinebench}.
For multilingual TBPR, we conduct experiments on their multilingual counterparts (\emph{i.e.}, CUHK-PEDES(M), ICFG-PEDES(M), and UFineBench(M)), recently proposed in~\cite{cao2025multilingual}. Details description of these datasets, evaluation metrics and implementation details are provided in the supplementary material.

\begin{table*}[t]
    \centering
    \small
    \caption{Ablation study on different pretraining datasets. "-" indicates no pretraining. Results are reported on CUHK-PEDES.}
    \label{tab: ablation_datasets}
    \setlength{\tabcolsep}{3pt}
    \renewcommand\arraystretch{1.2}
    \setlength{\extrarowheight}{3pt}

    \resizebox{0.92\linewidth}{!}{%
    \begin{tabular}{c|c|ccc|ccc|ccc|ccc|ccc}
        \hline\thickhline
        \rowcolor{lightgray}
        & & \multicolumn{3}{c|}{English} & \multicolumn{3}{c|}{Chinese} & \multicolumn{3}{c|}{French} & \multicolumn{3}{c|}{German} & \multicolumn{3}{c}{Avg} \\

        \rowcolor{lightgray}
         \multirow{-2}{*}{Method} & \multirow{-2}{*}{Pretraining Dataset} & R@1 & R@10 & mAP & R@1 & R@10 & mAP & R@1 & R@10 & mAP & R@1 & R@10 & mAP & R@1 & R@10 & mAP \\
        \hline\hline

        \multirow{3}{*}{Bi-IRRA~\cite{cao2025multilingual}} & - & 78.82 & 95.47 & 69.68 & 76.43 & 94.79 & 67.79 & 76.46 & 94.35 & 67.26 & 75.57 & 94.31 & 67.07 & 76.82 & 94.73 & 67.95 \\
         & LUPerson-M~\cite{tan2024harnessing} & 79.43 & 95.68 & 70.51 & 76.55 & 94.49 & 68.05 & 76.60 & 94.31 & 67.65 & 75.79 & 94.38 & 67.89 & 77.09 & 94.72 & 68.53 \\
         & MultiPedes & 79.64 & 95.55 & 70.60 & 76.86 & 94.57 & 68.56 & 76.77 & 94.54 & 68.38 & 76.68 & 94.57 & 68.05 & 77.49 & 94.81 & 68.90 \\ \hline
        \multirow{3}{*}{ViSeLA} & - & 78.65 & 95.21 & 70.33 & 76.08 & 94.17 & 68.30 & 76.65 & 94.23 & 68.35 & 75.92 & 94.18 & 67.87 & 76.83 & 94.45 & 68.71 \\
         & LUPerson-M~\cite{tan2024harnessing} & 78.95 & 95.13 & 70.76 & 76.36 & 94.20 & 68.57 & 76.89 & 94.22 & 68.54 & 76.23 & 93.86 & 68.41 & 77.11 & 94.35 & 69.07 \\
         & MultiPedes & 79.37 & 95.11 & 71.68 & 77.29 & 93.94 & 69.72 & 77.66 & 94.10 & 69.71 & 77.45 & 94.02 & 69.53 & 77.94 & 94.29 & 70.16  \\

        \hline\thickhline
    \end{tabular}%
    }
    \label{tab:ablation_datasets}
\end{table*}
\begin{table}[t]
    \centering
    \small
    \caption{Ablation study on key components of the proposed ViSeLA framework. S-MoE, V-MoE and L-MoE denote our proposed semantic-aware MoE, vanilla MoE, and language-aware MoE, respectively. Results are averaged over the CUHK-PEDES dataset. Results are averaged on CUHK-PEDES.}
    \label{tab:ablation}

    \renewcommand\arraystretch{1.2}
    \setlength{\extrarowheight}{3pt}

    \resizebox{\linewidth}{!}{%
    \begin{tabular}{c|c|ccc|cccc}
        \hline\thickhline
        \rowcolor{lightgray}      
         &  & \multicolumn{3}{c|}{Se-L} &  &  &  & \\
        \rowcolor{lightgray}
        \multirow{-2}{*}{No.} & \multirow{-2}{*}{Vi-A} & S-MoE & V-MoE & L-MoE & \multirow{-2}{*}{R@1} & \multirow{-2}{*}{R@5} & \multirow{-2}{*}{R@10} & \multirow{-2}{*}{mAP} \\
        \hline\hline
        1 &  & $\checkmark$ &  &  & 77.67 & 90.97 & 94.39 & 69.54 \\
        2 & $\checkmark$ &  &  &  & 77.21 & 90.73 & 94.20 & 69.56 \\ 
        3 & $\checkmark$ &  &  & $\checkmark$ & 77.49 & 91.00 & 94.36 & 69.78 \\ 
        4 & $\checkmark$ &  & $\checkmark$  &  & 77.37 & 90.91 & 94.42 & 69.61 \\ 
        \hline\rowcolor[HTML]{D7F6FF}
        5 & $\checkmark$ & $\checkmark$ &  &  & \textbf{77.94} & \textbf{90.94} & \textbf{94.29} & \textbf{70.16} \\

        \hline\thickhline
    \end{tabular}%
    }
\end{table}

\subsection{Comparisons with SOTA Methods}
In this section, we comprehensively evaluate the proposed ViSeLA against state-of-the-art methods under both multilingual TBPR and traditional English-only TBPR tasks.

\subsubsection{Comparison on Multilingual TBPR}

Table~\ref{tab:multilingual tbpr} reports the results on CUHK-PEDES(M), ICFG-PEDES(M), and UFineBench(M).
Given the substantial gains from large-scale pedestrian-centric pretraining, we categorize methods based on whether such pretraining is employed—as reported in their original papers—and perform comparisons within each group to ensure a fair evaluation.

Under both settings, ViSeLA consistently achieves the best performance across all datasets.
Specifically, without pretraining, ViSeLA outperforms Bi-IRRA by \(0.01/0.76\), \(1.23/1.84\), and \(0.64/0.44\) in average R@1/mAP across languages on CUHK-PEDES(M), ICFG-PEDES(M), and UFineBench(M), respectively.
With large-scale pretraining, the performance margins are further enlarged to \(0.85/1.63\), \(1.29/1.83\), and \(1.22/1.27\), demonstrating the strong effectiveness and robustness of ViSeLA in multilingual scenarios.

\subsubsection{Comparison on English-only TBPR}

Although our proposed ViSeLA targets multilingual TBPR, it naturally supports English-only inputs under the traditional setting. We evaluate it in this setting against state-of-the-art English-only TBPR methods.
Following prior practice, the methods are grouped based on whether they leverage large-scale pedestrian-centric pretraining (as reported in their original papers).

As shown in Table~\ref{tab:english_tbpr}, ViSeLA consistently achieves the best performance with and without pretraining, showing its robustness across different settings.  
Specifically, without pretraining, ViSeLA outperforms the current best method, RASA~\cite{bai2023rasa}, by \(1.11/0.08\) in terms of R@1/mAP on CUHK-PEDES and also surpasses the state-of-the-art method RDE~\cite{qin2024noisy} on ICFG-PEDES and CFAM~\cite{zuo2024ufinebench} on UFineBench.
With pretraining, it further surpasses HAM~\cite{jiang2025modeling}, which is pretrained on HAM-PEDES, by \(1.43/1.81\) in terms of R@1/mAP on CUHK-PEDES.
This improvement is well-motivated: the multilingual training corpus provides richer semantic diversity, and the joint alignment across languages and modalities implicitly enhances the cross-modal alignment, thereby boosting performance on English-only TBPR.

\subsection{Ablation Study}

\subsubsection{Effectiveness of MultiPedes}

To validate the effectiveness of our proposed pretraining dataset, we train both Bi-IRRA and ViSeLA under three settings: (i) without pretraining, (ii) with pretraining on LUPerson-M~\cite{tan2024harnessing}, and (iii) with pretraining on MultiPedes.

Results are reported in Table~\ref{tab: ablation_datasets}, from which we draw two key observations.  
First, across both architectures, pretraining on MultiPedes consistently yields higher average retrieval performance than pretraining on LUPerson-M. This confirms that MultiPedes provides more effective supervision for multilingual feature learning.  
Second, pretraining on LUPerson-M leads to imbalanced performance gains across languages for Bi-IRRA, whereas ViSeLA exhibits consistently balanced improvements. Specifically, Bi-IRRA achieves a notable gain of \(+0.61\) R@1 on English, but only marginal improvements on Chinese \((+0.12)\), French \((+0.14)\), and German \((+0.22)\). In contrast, ViSeLA obtains gains of \(+0.30\), \(+0.28\), \(+0.24\), and \(+0.31\) in R@1 on English, Chinese, French, and German, respectively—demonstrating more uniform cross-lingual transfer.
This discrepancy stems from the fact that LUPerson-M contains annotations exclusively in English, already introducing a language resource imbalance. The English-centric alignment mechanism in Bi-IRRA amplifies this bias during pretraining, while the vision-grounded multimodal alignment in ViSeLA anchors all languages to shared visual semantics, thereby mitigating English dominance and promoting balanced multilingual performance.

\subsubsection{Ablation on ViSeLA}
We conduct ablation studies to validate the effectiveness of the two key components in ViSeLA, namely the Vi-A and the Se-L module in Table~\ref{tab:ablation}.

\textbf{Effectiveness of Vi-A.}
Removing the Vi-A module (No.1 \emph{vs.} No.5) leads to the performance drop, confirming its effectiveness.  
Vi-A serves two critical functions in ViSeLA:  
(i) it extracts and aligns fine-grained semantic cues across visual and textual modalities, (ii) it leverages image semantics to anchor shared semantic concepts from multilingual texts, which guides the semantic-aware routing in Se-L.
In the absence of Vi-A, the model resorts to coarse global alignment alone, which not only weakens cross-modal alignment but also undermines the effectiveness of semantic-aware expert routing, ultimately degrading performance.

\textbf{Effectiveness of Se-L.}
Similarly, removing the Se-L module (No.2 \emph{vs.} No.5)—which incorporates a semantic-aware MoE (S-MoE)— also leads to the performance drop, confirming its effectiveness.  
Se-L assigns semantically similar tokens from different languages to the same expert, enabling more expressive modeling and facilitating the learning of language-invariant features.
Without Se-L, the model struggles to disentangle shared semantics across languages, leading to degraded performance.

Additionally, Se-L is built upon S-MoE, we compare it against two alternative MoE designs—vanilla MoE (V-MoE) and language-aware MoE (L-MoE)—to further validate its design.
V-MoE employs a simple linear router for expert assignment, while L-MoE routes tokens solely based on language identity, assigning all tokens from the same language to a shared expert.
First, comparing V-MoE with S-MoE (No.4 \emph{vs.} No.5) reveals a performance gap, indicating that increasing model capacity via the vanilla MoE is insufficient for multilingual TBPR.  
Semantic-aware routing is essential to effectively leverage the MoE architecture in this setting.
Second, S-MoE also consistently outperforms L-MoE (No.3 \emph{vs.} No.5), demonstrating that routing based on semantic similarity is more effective than language-based assignment.  
These results underscore a key insight: effective multilingual modeling requires disentangling and specializing shared semantic concepts across languages, rather than treating each language independently.

Furthermore, we provide more comprehensive experiments in the supplementary material.

\section{Conclusion}
We present MultiPedes, the first large-scale multilingual pretraining dataset for TBPR, comprising 10M image–text pairs to support large-scale pretraining. To ensure data quality, we design a multilingual quality-aware data construction pipeline. In addition, we propose ViSeLA, a vision-grounded and semantic-aware framework for multilingual retrieval. By jointly addressing pretraining data scarcity and model limitations, ViSeLA enables effective language-invariant representation learning and mitigates English-centric alignment bias. Extensive experiments on both multilingual and English-only benchmarks demonstrate the superiority of our framework. We expect this work to provide a strong foundation for future research in multilingual TBPR.

\begin{acks}
This work is supported by the National Natural Science Foundation of China under Grants 62476188.
\end{acks}

\bibliographystyle{ACM-Reference-Format}
\bibliography{references}

\end{document}